\documentclass[letterpaper]{article} 
\usepackage{aaai25}  
\usepackage{times}  
\usepackage{helvet}  
\usepackage{courier}  
\usepackage[hyphens]{url}  
\usepackage{graphicx} 
\usepackage{natbib}  
\usepackage{caption} 
\usepackage{algorithm}
\usepackage{algorithmic}

\usepackage{hhline}
\usepackage{multirow}
\usepackage{makecell}
\usepackage{xcolor}
\usepackage{tikz}
\usepackage{booktabs}
\usepackage{graphicx}
\usetikzlibrary{shapes, arrows, positioning}
\usetikzlibrary{positioning, shapes.geometric, arrows.meta}

\usepackage{newfloat}
\usepackage{listings}
\DeclareCaptionStyle{ruled}{labelfont=normalfont,labelsep=colon,strut=off} 
\floatstyle{ruled}
\newfloat{listing}{tb}{lst}{}
\floatname{listing}{Listing}
\nocopyright 

\usepackage{amsmath,amssymb,mathtools}

\usepackage[capitalize]{cleveref}
\crefname{figure}{Fig.}{Figures}
\Crefname{figure}{Figure}{Figures}


\usepackage{xfrac}
\usepackage{todonotes}
\makeatletter
\define@key{todonotes}{EK}[]{%
    \setkeys{todonotes}{author=EK,color=green}}%
\makeatother        

\title{Learn2Aggregate: Supervised Generation of Chv\'atal-Gomory Cuts Using Graph Neural Networks}
\author{
    Arnaud Deza\textsuperscript{\rm 1},
    Elias B. Khalil\textsuperscript{\rm 1},
    Zhenan Fan\textsuperscript{\rm 2},
    Zirui Zhou\textsuperscript{\rm 2},
    Yong Zhang\textsuperscript{\rm 2},
}
\affiliations{
    \textsuperscript{\rm 1}Department of Mechanical \& Industrial Engineering, University of Toronto\\
    \textsuperscript{\rm 2} Huawei Technologies Canada\\

}

\begin{document}

\maketitle

\begin{abstract}
We present \textit{Learn2Aggregate}, a machine learning (ML) framework for optimizing the generation of Chv\'atal-Gomory (CG) cuts in mixed integer linear programming (MILP). The framework trains a graph neural network to classify useful constraints for aggregation in CG cut generation. The ML-driven CG separator selectively focuses on a small set of impactful constraints, improving runtimes without compromising the strength of the generated cuts. Key to our approach is the formulation of a constraint classification task which favours sparse aggregation of constraints, consistent with empirical findings. This, in conjunction with a careful constraint labeling scheme and a hybrid of deep learning and feature engineering, results in enhanced CG cut generation across five diverse MILP benchmarks. 
On the largest test sets, our method closes roughly \textit{twice} as much of the integrality gap as the standard CG method while running 40\% faster. This performance improvement is due to our method eliminating 75\% of the constraints prior to aggregation.  
\end{abstract}

%

\section{Introduction}
Mixed integer linear programming (MILP) is an established mathematical optimization framework used in many industrial applications such as operation room scheduling, supply chain, and transportation. The integrality requirement makes MILP NP-hard in general and thus highly challenging to solve exactly in practice. Cutting planes (or cuts) play a pivotal role in strengthening the linear programming (LP) relaxation of MILPs, leading to tighter bounds on the optimal integer value. Cuts can be separated heuristically or by optimizing an appropriate measure of cut quality. For instance, Gomory Mixed-Integer (GMI) cuts are read from the simplex tableau of the LP relaxation; a simple, fast heuristic which may nonetheless generate weak cuts. 

In contrast, optimization-based separation, such as that of Chv\'atal-Gomory (CG) cuts, produces strong cuts but requires solving challenging auxiliary  MILPs. A CG cut is a weighted aggregation of a subset of the original constraints followed by a rounding down of the resulting coefficients. However, optimizing over the extremely large set of possible aggregation weights for all constraints becomes computationally prohibitive for large MILPs. Although screening rules to eliminate constraints for the special case of $\{0, \sfrac{1}{2}\}$-CG cuts \cite{zerohalfcuts}, they have not been explored for general CG cuts.

This paper bridges the gap between heuristic and optimization-based cut generation by using machine learning (ML) to accelerate the optimization of aggregation-based cuts. We address the computational bottleneck of cut optimization by leveraging a model trained to screen constraints, eliminating unnecessary ones from consideration for aggregation. This allows us to solve a \textit{reduced} separation problem over a much smaller subset of constraints and can be seen as a heuristic version of an exact optimization formulation for cut separation.
\begin{figure}[ht]
    \centering
    \includegraphics[scale=0.71]{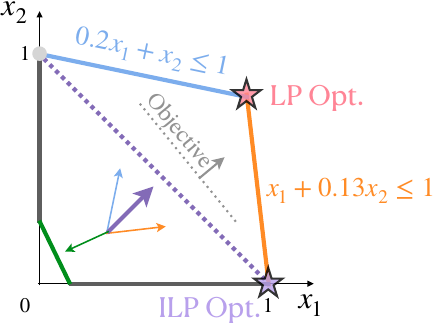}
    \caption{A binary two-variable problem. Constraints are solid and their associated normal vectors (excluding the bound constraints) are drawn, along with the maximization objective vector. Point ``LP Opt.'' is the solution of the LP relaxation whereas ``ILP Opt.'' is the desired integer optimum. The dashed cut, $x_1+x_2\leq 1$ in purple, is a strong cut which we would like to derive by aggregating the (normal vectors of the) dashed constraints; this cut tightens the LP such that its optimum is ``ILP Opt.''. The normal vector of the desired cut is in the cone of the blue and orange constraints and thus can be derived by aggregating them with appropriate weights (e.g., multiplying both of them with 0.9, adding them up, then rounding down the resulting vector; this is a CG cut). The green constraint is not useful and can be excluded from consideration when aggregating.
    }
    \label{fig:illustration}
\end{figure}

Our method involves a careful combination of the following empirical observations and technical ideas:
\begin{enumerate}
    \item On five families of MILP problems (facility location, packing, stable set, p-median, set covering), we observe that CG cuts generated via optimization~\cite{lodi} typically aggregate a tiny fraction of all constraints (\cref{tab:dataset_stats}). If those constraints could be identified \textit{in advance of cut generation}, the optimization would likely be much faster.~\Cref{fig:illustration} illustrates the idea.
    \item CG optimization can produce many cuts of varying quality. We argue for learning only from non-dominated cuts, i.e., cuts that are binding at the LP optimum after they are added. This results in better labels for training a constraint classifier.
    \item Our constraint classifier is a hybrid of feature engineering --- representing constraints and variables using metrics established in the mathematical optimization literature --- and graph neural networks (GNNs). This design choice makes learning the GNN far easier than using a pure deep learning approach.
\end{enumerate}

We evaluate the ML-based separator against heuristic CG cuts and exact separation on the five families of problems in terms of final integrality gap closure (IGC, i.e., how close the LP relaxation bound is to the optimal integer value) and total cut generation time. We find that:
\begin{enumerate}
    \item On average, we remove 75\% of the constraints, leading to mean IGC improvements of 2\%, 23\%, and 93\% at time reductions of 57\%, 64\%, and 41\% for small, medium, and large instances, respectively. These are substantial improvements that validate our motivating observation 1. above as well as the GNN design described in 3.

    \item We show that learning only from non-dominated cuts (2. above) is strictly better than the more straightforward approach of learning from all CG cuts.

    \item A GNN trained on small instances of a given problem performs well on medium and large instances. This can save substantial data collection and training time which can be constrained in some applications.
\end{enumerate}

\section{Background}
\label{sec:ch2}
\subsection{Integer Linear Programming (ILP)}
An ILP is an optimization problem involving $n$ discrete variables $\boldsymbol{x} \in \mathbb{Z}^{n}$ that aims to minimize a linear objective function $\boldsymbol{c}^{\intercal}\boldsymbol{x}$ over $m$ linear constraints $ \boldsymbol{A}\boldsymbol{x}\leq \boldsymbol{b}$. Assuming the variables are nonnegative, an ILP reads as:
\begin{equation}
    \label{eqn:ILP}
    z^{IP} = \text{min}\{\boldsymbol{c}^{\intercal}\boldsymbol{x} |  \boldsymbol{A}\boldsymbol{x}\leq \boldsymbol{b},   \boldsymbol{x} \in \mathbb{Z}^{n}_{+}\}
\end{equation}
This formulation induces two key polyhedra: the continuous polyhedron $P$ and its integer hull $P_I$, defined as:
\begin{align}
\label{eqn:LP}
    P &:= \{\boldsymbol{x}\in\mathbb{R}^{n}_+ : \boldsymbol{A}\boldsymbol{x}\leq \boldsymbol{b}\}\\
    P_I &:=  \rm{conv}\{\boldsymbol{x}\in\mathbb{Z}^{n}_{+} : \boldsymbol{A}\boldsymbol{x}\leq \boldsymbol{b}\} = \rm{conv}(P\cap \mathbb{Z}^{n}_{+})
\end{align}

The polyhedron $P$ represents the feasible region of the LP relaxation, while $P_I$ is the convex hull of feasible integer points. Solving ILPs is challenging given their NP-hard nature, but modern solvers effectively tackle large instances using a combination of exact and heuristic methods. Solvers mainly rely on \emph{Branch and Bound} (B\&B) \cite{b_and_b}, a tree search algorithm that iteratively solves LP relaxations, tightens bounds by branching on fractional variables, prunes suboptimal branches, and uses cuts to further constrain $P$ and approximate $P_I$. Here, we focus on how to get strong cuts quickly.

\subsection{Cutting Planes} 
\label{sec:num2}
Cutting planes are valid linear inequalities to problem~\eqref{eqn:ILP} of the form $\mathbf{\boldsymbol{\alpha}}^T \boldsymbol{x} \leq \alpha_0, \mathbf{\boldsymbol{\alpha}} \in \mathbb{R}^n, \alpha_0 \in \mathbb{R}$. They are ``valid" in the sense that adding them to the polyhedra $P$ is guaranteed not to cut off any feasible solutions to $P_I$. Additionally, one seeks cuts that separate the current LP solution, $\boldsymbol{x}^*_{LP}$, from $P_I$. Although adding more cuts can help achieve tighter relaxations in principle, a clear trade-off exists: as more cuts are added, the size of the LP relaxation grows resulting in an increased cost in LP solving at the nodes of the B\&B tree~\cite{Tobias_thesis}. Adding too few cuts, however, may lead to a large number of nodes in the search tree as more branching is required.

A typical ``cutting loop'' consists of a number of rounds in which a new LP optimum is separated by adding one or more cuts until an integer LP optimum is found (indicating convergence to an optimal solution) or other termination criteria are met. Despite theoretical finite convergence results for the cutting plane method using Gomory cuts, numerical issues will often prevent convergence to an optimal solution in practice. It is thus typical to cut in rounds until the LP has been tightened sufficiently. 

A standard metric to evaluate the cutting plane method is the \emph{integrality gap} (IG). Let $z^t\in\mathbb{R}$ be the objective value of the LP after $t$ rounds of cutting and let the bound difference $g^t \coloneqq z^{IP} -z^{t}\geq 0$. The integrality gap closure (IGC) \cite{Columbia} is measured as :
\begin{equation}
    IGC^{t}\coloneqq 100 \cdot \frac{g^0-g^t}{g^0}= 100 \cdot \frac{z^t-z^0}{z^{IP}-z^0} \in [0,100]
    \label{eq:igc}
\end{equation}
IGC represents the percentage of the integrality gap that has been closed after $t$ separation rounds.

\subsection{General Valid Inequalities: Chv\'atal-Gomory Cuts}
This work focuses on the widely used CG cuts~\cite{chv} which are generated by aggregating a subset of the original constraints of $P$ followed by appropriate rounding. CG cuts are defined as the following type of valid inequality for $P_I$:
\begin{equation}\label{eq:cg}
\begin{aligned}
    \boldsymbol{\alpha}^\intercal\boldsymbol{x} \leq \alpha_0, \text{with }  
    \boldsymbol{\alpha}=\lfloor \boldsymbol{u}^\intercal \boldsymbol{A} \rfloor\in\mathbb{Z}^{n} , \alpha_0=\lfloor \boldsymbol{u}^\intercal\boldsymbol{b}\rfloor\in\mathbb{Z}
\end{aligned}
\end{equation}
The entries of $\boldsymbol{u}$ are the nonnegative CG aggregation coefficients, one for each constraint. $\boldsymbol{\alpha}$ are the integer cut coefficients and $\alpha_0$ is the right-hand side coefficient of the cut. In line with prior research on CG cuts~\cite{lodi}, we focus on rank-1 cuts, i.e, cuts that only depend on the constraints defining $P$.

\subsection{Chv\'atal-Gomory Cut Generation: CG-MIP}
It is possible to generate a CG cut heuristically by reading the aggregation coefficients from the simplex tableau of the LP relaxation. However,~\citet{lodi} have shown that stronger cuts with higher IGC can be generated if one~\textit{optimizes} for an appropriate objective function.
Concretely, to separate a point $\boldsymbol{x}^*$,~\citet{lodi} propose an auxiliary MILP for CG cut separation, the CG-MIP in~\eqref{eq:cgmip}. Here, $J(\boldsymbol{x}^*)=\{j\in \{1,\cdots,n\}: x_j^*>0\}$, denotes the support of $\boldsymbol{x}^*$ and $\delta$ is a small user-defined parameter (we use $\delta = 0.01$).
\begin{align}
\label{eq:cgmip}
\max  & \sum_{j\in J(\mathbf{x^*})} \alpha_j x^*_j - \alpha_0 &  \\
\mbox{subject to }   &f_0 = \boldsymbol{u}^\intercal\boldsymbol{b} -  \alpha_0 \;\; & \notag\\
                    &f_j = \boldsymbol{u}^\intercal\boldsymbol{A_j} - \alpha_j \;\; & \forall j \in J(\boldsymbol{x^*})\notag\\
                   &0 \leq f_j \leq 1-\delta \;\; & \forall j \in J(\boldsymbol{x^*})\cup \{0\}\notag\\
                   &0 \leq u_i \leq 1-\delta \;\; & \forall i\in\{1,\dots, m\}\notag\\
                   &\alpha_j \in \mathbb{Z} \;\; & \forall j \in J(\boldsymbol{x^*})\cup \{0\}\notag\\
                   &\sum\limits_{j\in J(\boldsymbol{x^*})} \alpha_j x^*_j - \alpha_0 \geq 0.01 \;\; & \notag
\end{align}

As in all MILP-based separators, CG-MIP's objective maximizes cut violation which must be at least 0.01. CG-MIP has $\mathcal{O}(n)$ integer and $\mathcal{O}(n+m)$ continuous variables, making its size linear in the original ILP's. The constraints implement~\eqref{eq:cg}: the aggregation of the inequalities using the variable weights $\alpha_j$ and their rounding down. CG-MIP can return a set of cuts using the off-the-shelf capability of modern MILP solvers to collect multiple feasible solutions (the solution pool). While effective, CG-MIP's reliance on solving a MILP makes it computationally expensive, especially in the B\&B tree, leading to its default deactivation in modern solvers (e.g., in SCIP~\cite{scip8}). 
Note that CG-MIP applies only to pure ILP but can be extended to MILP by the projected Chv\`atal-Gomory cuts of~\citet{Bonami2008}. More details of this simple extension, which we use for the facility location problem, are left to~\cref{app:procg}. 

\section{Related Work}\label{sec:ch3}
\paragraph{ML for MILP}\citeauthor{bengio} (2021) survey recent successful ML approaches to automate decision-making in MILP solvers through learned models. Notable examples include the learning of computationally challenging variable selection rules for B\&B \cite{khalil2016learning,GCNN,seyfi2023exactcombinatorialoptimizationtemporoattentional}, learning to schedule heuristics \cite{learn_to_schedule} or estimating variable values \cite{nair2021solvingmixedintegerprograms,mip_gnn}.

\paragraph{Representing MILPs for ML} In recent years, graph neural networks (GNNs) have emerged as a popular architecture for several ML applications for MILP~\cite{cappart}. GNNs can handle sparse MILP instances and are permutation-invariant, making them well-suited for representing MILP instances. The GNN operates on the so-called \textit{variable-constraint graph} (VCG) of a MILP, first introduced by \citet{GCNN}. The VCG has $n$ variable nodes and $m$ constraint nodes corresponding to the decision variables and constraints of~\eqref{eqn:ILP}. Edges between a variable node $j$ and constraint node $k$ represent the presence of variable $x_j$ in constraint $k$ whenever the weight of the edge, $A_{jk}$, is nonzero.

\paragraph{ML for cutting planes tasks}  Recent studies demonstrate success in integrating ML into cutting plane subroutines such as cut selection \cite{Columbia,look_ahead,ACS}, cut addition \cite{local_cuts}, and cut removal \cite{puigdemont2024learningremovecutsinteger}. For a comprehensive survey on this topic, we refer the reader to \citet{Deza_2023}.

These approaches are orthogonal to ours: they rely on existing (heuristic) cut generators and seek only to select from those generators' cuts. The gap closed by a cut selection strategy is inherently upper-bounded by that of the whole set of cuts being considered. For example, while \citet{Columbia} successfully applied ML to select Gomory cuts from the simplex tableau, these weak, un-optimized cuts limited their experiments to very small instances. In contrast, we focus on larger instances where exact separation is challenging but produces stronger cuts. As a point of comparison, our smallest packing instances have 100 variables and constraints whereas the largest considered by~\citet{Columbia} have 60; our largest have 500, roughly 8 times more.

\paragraph{ML for cutting plane generation} Despite this success, little attention has been paid to ML for cut \textit{generation}, with only two papers to our knowledge. \citet{chételat2022continuous} frame cut generation as a continuous optimization problem over weights parameterizing families of valid inequalities (GMI cuts) that are optimized via gradient descent to maximize the dual bound of the LP relaxation with the generated cuts. Empirical results demonstrate improved dual bounds over classical GMI cuts read from the simplex tableau, although at a higher computational cost. This method applies on a per-instance basis and is not designed for the typical ML setting of learning over a distribution of instances. 

In a similar vein, \citet{dragotto2023differentiable} train a recurrent neural network to dynamically adjust the split-cut separator parameters (parametric LP) from \citet{Balas2008}. This framework shows good in-distribution generalization and effectively reduces integrality gaps, albeit at a significant computational overhead. This is due to the use of cvxpylayers \cite{cvxpylayers2019}, which does not leverage sparse linear algebra, requiring recomputation of objects during both forward/backward passes. This restricts their experiments to small-scale MILP instances and very few cuts generated, making cut generation significantly slower than solving the parametric LP of~\citet{Balas2008}.

Both approaches highlight the potential of ML-driven cut generation but come with increased computational complexity, making them significantly slower than traditional methods, a bottleneck which we seek to overcome here.

\section{Methodology}\label{sec:ch4}
We propose \textit{Learn2Aggregate}, a binary classification framework to identify useful constraints for aggregation, thereby reducing the computational complexity of an exact separator for a family of aggregation-based cuts, here CG cuts.

\subsection{ML-Driven Constraint Aggregation}
At each iteration $t$ of the cutting plane method, the state $S_t$ of the algorithm is represented by the tuple $\{\mathbf{A}, \mathbf{b}, \mathbf{x}^{t}_{LP}, U\}$ where $\mathbf{x}^{t}_{LP}$ denotes the current (fractional) LP relaxation solution and the set $U$ represents the $k$ CG cuts generated thus far as characterized by their aggregation coefficients $U = \{ \mathbf{u}_1, \mathbf{u}_2, \ldots, \mathbf{u}_k \}$.
We consider CG-MIP to be our ``oracle'' for good cuts and use it in this cutting loop to generate the set $U$. The binary classification target $y_i$ of a constraint $i$ is defined based on its participation in the formation of any CG cut. Specifically, $y_i$ is given by:

\begin{equation}
\label{label_curation}
y_i = \mathbb{I}\{\exists j \in \{1, 2, \ldots, k\} \text{ s.t. } u_{ji} > 0\},
\end{equation}
where $u_{ji}$ is the $i$-th coefficient of the $j$-th aggregation vector $\mathbf{u}_j$ and $\mathbb{I}$ is the indicator function. In words, the binary label $y_i$ is 1 iff the constraint $i$ has a nonzero coefficient in one or more of the aggregation vectors $U$.~\Cref{tab:dataset_stats} shows that for the representative set of MILPs we consider, the sparsity of the aggregation vectors $\mathbf{u}$ of the cuts produced by CG-MIP is high, motivating our constraint classification approach.

\subsection{Instance-Level Constraint Classification}
Constraint classification can be done at the constraint or instance level. Constraint-level prediction involves a model that predicts whether a constraint should be utilized based on a feature vector that represents it. Instance-level prediction involves training a GNN that operates on the VCG representation of a MILP where the model simultaneously classifies all constraints. The instance-level approach has a better chance at capturing the complex interdependencies between variables and constraints than a local constraint-level one. Due to space limitations, we only present results for instance-level prediction. 

We model the CG cut separation problem at iteration $t$ using the VCG representation $G=\{\mathcal{C}, \mathcal{V},\mathcal{E}\}$ of the MILP instance, where $\mathcal{C}$ and $\mathcal{V}$ denote sets of constraints and variables, respectively, and $\mathcal{E}$ represents the edges between them. The constraint and variable feature matrices are represented by $\mathbf{C} \in \mathbb{R}^{m \times 53}$ and $\mathbf{V} \in \mathbb{R}^{n \times 18}$, respectively. Those features capture the relationship between the fractional LP solution being separated, $\mathbf{x}^{t}_{LP}$, and the constraints. Matrix $\mathbf{E} \in \mathbb{R}^{(m \cdot n) \times 1}$ contains edge features. A complete description of all features is included in~\cref{sec:data_feats}.

\subsection{Integrating Efficient Cut Selection into Generation}
\label{sec:efficientcutselection}
Our goal is to learn to select constraints that lead to cuts maximizing dual-bound improvement (i.e., the IGC defined in~\eqref{eq:igc}) rather than maximizing violation, a simple linear surrogate of cut quality.
At round $t$, CG-MIP generates $k$ cuts to separate the current fractional solution $\mathbf{x}^{t}_{LP}$. Which of these cuts is most useful in tightening the LP? We can find out by adding all $k$ cuts to the current relaxation and solving the new LP to obtain the relaxed solution $\mathbf{x}^{t+1}_{LP}$. We argue that cuts that are tight to $\mathbf{x}^{t+1}_{LP}$, i.e., for which $\mathbf{\alpha}^\intercal \mathbf{x}^{t+1}_{LP}=\alpha_0$, dominate the other cuts. As such, classification labels $y_i$ for round $t$ are generated based only on cuts these tight cuts, rather than all $k$ cuts. Only the deepest cuts are used to label the constraints, resulting in a refined signal for training.

\subsection{Training and Neural Network Architecture}
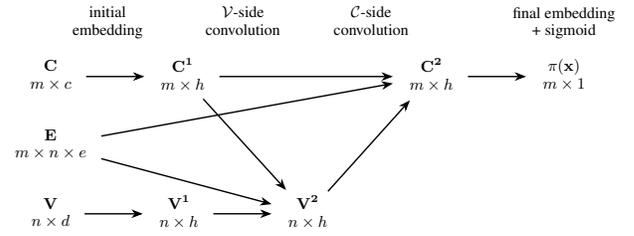
\begin{figure}[h!]

  \centering
\resizebox{\columnwidth}{!}{%
  \begin{tikzpicture}[
    node distance=0.5cm and 0.5cm,
    every node/.style={font=\small, inner sep=2pt},
    >={Stealth},
    shorten >=1pt,
    auto,
    thick,
    scale=0.04
  ]

  \node (C) {\begin{tabular}{c} $\mathbf{C}$ \\ $m \times c$ \end{tabular}};
  \node (E) [below=of C] {\begin{tabular}{c} $\mathbf{E}$ \\ $m \times n \times e$ \end{tabular}};
  \node (V) [below=of E] {\begin{tabular}{c} $\mathbf{V}$ \\ $n \times d$ \end{tabular}};
  \node (V1) [right=of V, xshift=0.7cm] {\begin{tabular}{c} $\mathbf{V^1}$ \\ $n \times h$ \end{tabular}};
  \node (C1) [right=of C, xshift=0.7cm] {\begin{tabular}{c} $\mathbf{C^1}$ \\ $m \times h$ \end{tabular}};
  \node (V2) [right=of V1, xshift=0.7cm] {\begin{tabular}{c} $\mathbf{V^2}$ \\ $n \times h$ \end{tabular}};
  \node (C2) [right=of C1, xshift=3.0cm] {\begin{tabular}{c} $\mathbf{C^2}$ \\ $m \times h$ \end{tabular}};
  \node (final) [right=of C2, xshift=0.7cm] {\begin{tabular}{c} $\pi(\mathbf{x})$ \\ $m \times 1$ \end{tabular}};

  \node[above=0.2cm of C1, xshift=-1.5cm] (init) {\begin{tabular}{c} initial \\ embedding \end{tabular}};
  \node[above=0.2cm of C1, xshift=1.2cm] (Cside) {\begin{tabular}{c} $\mathcal{V}$-side \\ convolution \end{tabular}};
  \node[above=0.2cm of C2, xshift=-1.2cm] (Vside) {\begin{tabular}{c} $\mathcal{C}$-side \\ convolution \end{tabular}};
  \node[above=0.2cm of final] (final_label) {\begin{tabular}{c} final embedding \\ + sigmoid \end{tabular}};

  \draw[->] (V) -- (V1);
  \draw[->] (E) -- (V2);
  \draw[->] (E) -- (C2);
  \draw[->] (C) -- (C1);
  \draw[->] (C1) -- (C2);
  \draw[->] (C1) -- (V2);
  \draw[->] (V1) -- (V2);
  \draw[->] (V2) -- (C2);
  \draw[->] (C2) -- (final);

  \end{tikzpicture}
  }
  \caption{Schematic of the constraint classification GNN with constraint node features $\boldsymbol{C}$, variable node features $\boldsymbol{V}$, and edge features $\boldsymbol{E}$. The embedding size is $h$.
  }
  \label{fig:embedding_convolutionf}
\end{figure}
We train the graph convolutional neural network proposed in \cite{GCNN}, denoted as $\pi$, with a slight modification in order to output logits $z_i$ for each constraint $i$. The GNN model architecture is depicted in ~\cref{fig:embedding_convolutionf}. The model initially embeds all input feature vectors $\mathbf{c}_i$ and $\mathbf{v}_i, \forall i \in \mathcal{C}, j \in \mathcal{V}$ to a common embedding size $h$, through a 2-layer perceptron. Then, the model performs one graph convolution involving two interleaved half-convolutions: (1) $\mathcal{V}$-side convolution from constraints to variables: 
$$\mathbf{v}_j \leftarrow \phi_{\mathcal{V}} \bigg(\mathbf{v}_j, \sum_{i:(i,j) \in \mathcal{E}} \psi_{\mathcal{V}} (\mathbf{c}_i, \mathbf{v}_j, e_{i,j}) \bigg), \forall j \in \mathcal{V},$$ 
and (2) $\mathcal{C}$-side convolution from variables to constraints: 
$$\mathbf{c}_i \leftarrow \phi_{\mathcal{C}} \bigg(\mathbf{c}_i, \sum_{j:(i,j) \in \mathcal{E}} \psi_{\mathcal{C}} (\mathbf{c}_i, \mathbf{v}_j, e_{i,j}) \bigg), \forall i \in \mathcal{C}.$$ 
In these equations, \( \phi_{\mathcal{C}}, \phi_{\mathcal{V}}, \psi_{\mathcal{C}}, \) and \( \psi_{\mathcal{V}} \) are 2-layer perceptrons with ReLU activation functions. To obtain $\pi$, we apply a final 2-layer perceptron on the constraint nodes $\mathbf{c}_i$ to produce logits $z_i$, which are then transformed into probability scores $\hat{y}_i$ using a sigmoid function: $\hat{y}_i = \sigma(z_i) = \frac{1}{1 + e^{-z_i}}$. The model is trained to minimize a weighted binary cross-entropy loss, accounting for the significant class imbalance where class-0 labels vastly outnumber class-1 labels. The loss for predictions $\mathbf{\hat{y}}$ of $m$ target values $\mathbf{{y}}$ is given by:
\begin{equation*}
\label{bce_loss}
\mathcal{L}(\mathbf{\hat{y}},\mathbf{{y}}) = -\frac{1}{m} \sum_{i=1}^{m} w_1 y_i \log(\hat{y}_i) + w_0 (1 - y_i) \log(1 - \hat{y}_i)
\end{equation*}
where $w_1$ and $w_0$ are weights assigned to class-1 and class-0, respectively, to counterbalance the skewed label distribution (see 'Sparsity' in~\cref{tab:dataset_stats}). 
A tunable classification threshold of \( \tau \in (0,1) \) is used to classify a constraint: $\tilde{y}_i = \mathbb{I}\{\hat{y}_i \geq \tau\}$.

\section{Experimental Setup}\label{sec:ch5}
Next, we describe our experimental setup. More detailed information can be found in the appendix. 

\begin{table}[ht]
\centering
\resizebox{\columnwidth}{!}{%
\begin{tabular}{cccc} 
\toprule
{\textbf{Problem}} & {\textbf{Size}} & \textbf{\#Vars / Cons}  & {\textbf{Sparsity (\%)}} \\
\midrule
\multirow{3}{*}{\textbf{Stable Set (SS)}}  & S & 50 / 203 & 85.82 \\
 & M & 75 / 535 & 90.43 \\
 & L & 100 / 1264 & 93.14 \\
\hline
\multirow{3}{*}{\textbf{P-Median (PM)}} & S & 110 / 32 & 70.59 \\
 &  M & 240 / 47 & 58.14  \\
 & L & 420 / 62 & 59.52  \\
\hline
\multirow{3}{*}{\textbf{Set Cover (SC)}} & S & 500 / 125 & 81.93 \\
 & M & 500 / 250 & 92.74 \\
 & L & 1000 / 250 & 98.03 \\
\hline
\multirow{3}{*}{\textbf{Binary Packing (BP)}} & S & 100 / 100 & 93.06 \\
 & M & 300 / 300 & 96.82 \\
 & L & 500 / 500 & 97.94 \\
\hline
\multirow{3}{*}{\textbf{Facility Location (FL)}}  & S & 2550 / 2601 & 97.44 \\
 & M & 5100 / 5151 & 98.91 \\
 & L & 10100 / 10201 & 99.36 \\
 \bottomrule
\end{tabular}
}
\caption{MILP benchmarks and their sizes: S, M, and L refer to small, medium, and large respectively.
}
\label{tab:dataset_stats}
\end{table}

\paragraph{Datasets} We experiment with five families of MILPs used in the research area of ML for MILP. They span a diverse range of combinatorial structures as can be seen in~\cref{tab:dataset_stats}. For each family of problems, three instance sizes are considered. For each family and size, a dataset of 200 instances is generated. We use MIPLearn \cite{miplearn} for the generation of PM and SS instances and Ecole \cite{prouvost2020ecolegymlikelibrarymachine} for the generation of FL and SC instances. Details regarding problem formulation and instance generation is also included in~\cref{sec:synthetic_MILP_instances}. To collect training data (on a random subset of the 200 instances), we deploy the full CG-MIP separator in a pure cutting plane method to collect features/labels as well as baseline performance for the full separator.~\Cref{tab:dataset_stats} summarizes the instance size statistics as well as their average sparsities as measured by the fraction of constraints in an instance that receive a label of zero based on the procedure described in the last paragraph of~\cref{sec:efficientcutselection}.

\paragraph{Baselines} We use two baselines to evaluate the proposed ML-driven  separator: (1) the full exact CG-MIP separator and (2) heuristic GMI cuts from the simplex tableau. The former can produce strong cuts at a high computational cost whereas the latter produces weak cuts extremely quickly.

\paragraph{Cutting Plane Setting} We evaluate our approach in a cutting plane method where CG-MIP is used to generate CG cuts at every round. As done by \citet{lodi}, GMI cuts read from the simplex tableau are added to the root LP relaxation prior to any CG cut generation. We restrict the maximum number of cut generation rounds to 100 and terminate early if IGC stagnates over 7 rounds to avoid diminishing returns. Additionally, motivated by the CG-MIP implementation in SCIP, we use a complemented MIR cut generation heuristic based on CG-MIP's solution pool.

\paragraph{CG-MIP Parameters} CG-MIP can run unnecessarily long without appropriate safeguards. CG-MIP's execution is stopped if an integer (optimal) solution has been found or if one of the following conditions are met: (1) A 15-second time limit is hit, (2) a feasible solution limit of 5000 is hit, or (3) an incumbent (i.e., improving) solution limit of 1000. The conservative time limit of 15 seconds is used to return cuts fast if any are found. Otherwise, the time limit is doubled and the solution limit is set to 1, i.e., the solver terminates on finding the first cut. The time limit is doubled until a total time limit of 180 seconds is reached, in which case the cutting plane method terminates. Additionally, to ensure CG-MIP focuses on finding feasible solutions rather than proving optimality, we set SCIP's internal parameter \textit{emphasis} to feasibility (\cref{app:impl_eval_details}). We use the same parameters when deploying the trained classifier to solve a reduced CG-MIP separator, ensuring a fair comparison.

\paragraph{Training Details} For each dataset, we use 100, 50, and 50 instances for training, validation, and testing. GNN hyperparameters are tuned using 80 random search trials. After training, the GNN's classification threshold (between zero and one) is tuned to maximize the F1-score on validation data. Tuning this threshold is crucial for striking a balance between accurately identifying useful constraints and avoiding the risk of over-constraining CG-MIP, which could lead to infeasibility in the cut generation process. Models are trained with Pytorch 1.10 \cite{paszke2019pytorchimperativestylehighperformance} and SCIP 8.0.0 \cite{scip8} is used as the MIP solver. More details can be found in~\cref{table:hyperparams1,table:hyperparams2} in the appendix.

\paragraph{Evaluation Metrics} To compare the reduced and full CG-MIP separator, we focus on size reduction and end-to-end metrics. The former is measured by the average percentage of constraints excluded from aggregation based on the classifier's predictions. The latter focuses on the final IGC after successive cut generation rounds and cumulative CG-MIP runtime, recorded in seconds. For these, we report the mean IGC and time using shifted geometric means with shifts of 0.5 and 5, respectively, as is standard in MILP benchmarking~\cite{Tobias_thesis}; see~\cref{app:impl_eval_details} for a definition. Additionally, we compare both separators by calculating the ratio of the means of the reduced separator for IGC and time, to those of the full separator. We perform a win/loss/tie analysis where we compare the reduced separator against the full separator for each instance. A \textit{win} (or \textit{loss}) in time indicates that the reduced separator is 10\% faster (or slower), while a \textit{win} (or \textit{loss}) in IGC signifies a 1\% higher (or lower) final IGC. An \textit{absolute win} occurs when time favours the reduced separator whilst beating or tying in IGC, and an \textit{absolute loss} occurs when both metrics incur a loss.

\section{Experimental Results}\label{sec:ch6}
Does ML-driven constraint selection accelerate cut generation and close more gap? We answer this question before analyzing the effect of using only tight cuts for constraint labeling and our models' size generalization capabilities.

\begin{table}[ht]
\centering
\resizebox{\linewidth}{!}{
\begin{tabular}{cccccc}
\toprule
\multirow{2}{*}{\textbf{Dataset}} & \multirow{2}{*}{\textbf{Size}} & \multicolumn{2}{c}{\textbf{Ratio}} & \textbf{Abs. Perf} & \textbf{\% Size}  \\
    &   & \textbf{IGC} $\uparrow$ & \textbf{Time} $\downarrow$ &  \textbf{Win/Loss} & \textbf{Reduction} \\
\midrule
\multirow{3}{*}{SS} & S &  0.96 & \textbf{0.33}  & 84 / 2 & 63.50  \\
 & M &  \textbf{1.28} & \textbf{0.19}  & 80 / 2 & 67.99  \\
 & L &  \textbf{2.59} & \textbf{0.9}  & 48 / 4 & 72.53  \\
\hline
\multirow{3}{*}{PM} & S &  \textbf{1.0} &  \textbf{0.3} & 84 / 0 & 62.91  \\
 & M &  \textbf{1.0} & \textbf{0.56}  & 74 / 4 & 49.30  \\
 & L &  \textbf{1.0} & \textbf{0.65}  & 82 / 2 & 37.98  \\
\hline
\multirow{3}{*}{SC} & S &  0.97 & \textbf{0.47} &  80 / 0 & 68.33  \\
 & M &  \textbf{1.31} & \textbf{0.23}  & 78 / 4 & 67.14  \\
 & L &  \textbf{2.99} & \textbf{0.66}  & 60 / 6 & 70.05  \\
\hline
\multirow{3}{*}{BP} & S &  \textbf{1.01} & \textbf{0.49} &  84 / 6 & 90.95  \\
 & M &  \textbf{1.14} & \textbf{0.31} &  80 / 2 & 95.33  \\
 & L &  \textbf{1.45} & \textbf{0.37} &  76 / 8 & 96.55  \\
\hline
\multirow{3}{*}{FL} & S &  \textbf{1.14} & \textbf{0.54} &  28 / 4 & 92.80  \\
 & M &  \textbf{1.41} & \textbf{0.51} &  34 / 6 & 96.22  \\
 & L &  \textbf{1.63} & \textbf{0.39} &  50 / 2 & 97.18  \\
\hhline{======}
\multirow{3}{*}{\textbf{Average}} & \textbf{S} &  \textbf{1.02} & \textbf{0.43}  & 72 / 2 & 75.70  \\
 & \textbf{M} &  \textbf{1.23} & \textbf{0.36} &  69 / 4 & 75.20  \\
 & \textbf{L} &  \textbf{1.93} & \textbf{0.59}  & 63 / 4 & 74.86  \\
\bottomrule
\end{tabular}
}
\caption{Ratios of mean test IGC and time of the reduced separator to those of the full separator followed by the absolute profile of win/loss counts and CG-MIP size reduction. Ratios are bolded when the reduced separator outperforms the full separator. Higher IGC ratios and lower time ratios are better for our method, respectively.}
\label{table:end2end}
\end{table}

\paragraph{ML closes more gap in less time.}
\cref{table:end2end} shows, on unseen test instances, the ratio of the shifted geometric means between the reduced separator and the full separator for the final IGC, total cut generation time (seconds), and number of cutting plane rounds. It also includes the percentage of constraints removed by the reduced separator (\% Size Reduction) and the absolute win/loss counts. The results demonstrate that we are able to achieve significant reductions --- above 60\% size for all problems --- in the size of the CG-MIP separator while maintaining comparable or superior final IGC (most IGC ratios exceed 1). For instance, across binary packing datasets, the reduced separator achieves an average IGC improvement of 20\% with a 60\% reduction in cut generation time, while removing over 94\% of constraints. We observe a consistent trend across problems and sizes, on average removing 75\% of constraints while achieving a 2\%, 23\%, and 93\% mean IGC improvement at a 57\%, 64\%, and 41\% mean time improvement for small, medium, and large instances respectively. In the appendix, we include raw values for mean IGC and time in~\cref{tab:raw_end2end} as well as test accuracy and recall in~\cref{table:classification_results} .

\paragraph{Zooming in on the cutting loops.}
\begin{figure}[t]
    \centering
    \includegraphics[scale=0.29]{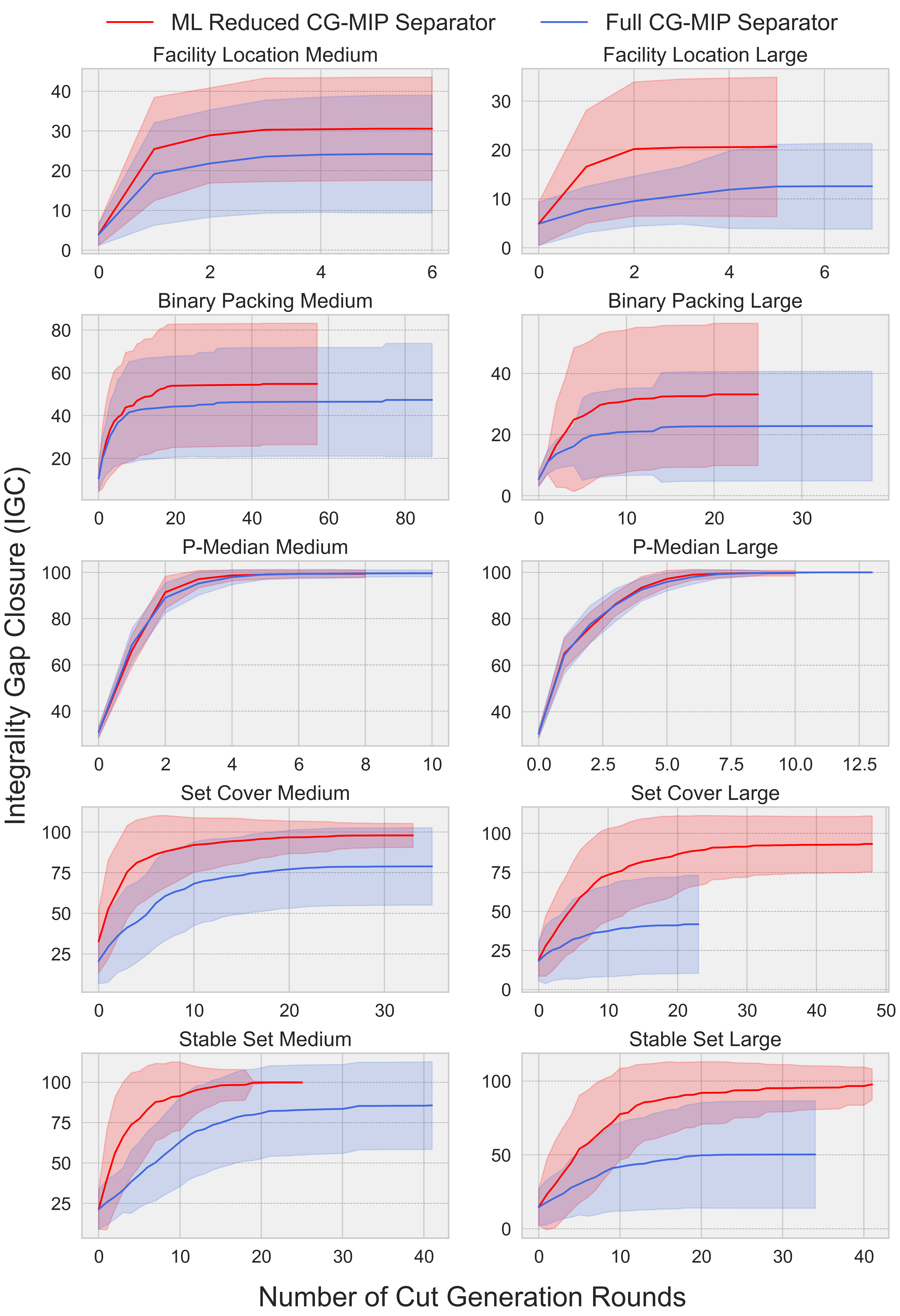}
    \caption{Mean IGC and standard deviation (shaded) v.s. round for medium/large test instances. The reduced and full CG-MIP separators are shown in red and blue, respectively.}
    \label{fig:igc_vs_round}
\end{figure}
\Cref{fig:igc_vs_round} shows the evolution of the primary evaluation metric, the IGC, over rounds of cutting by both the ML-reduced and full separators. Of note is that the former typically dominates the latter even very early on in the process. For most of these instance sets, our method typically terminates sooner than the full separator; this may explain the running time reductions observed in~\cref{table:end2end}. However, for large instances of Set Cover and Stable Set (last two bottom-right sub-plots), ML runs for more rounds. This may be explained by the hardness of these two problems at scale, while noting that they also exhibit the largest improvements in IGC: ratios of 2.59 and 2.99, respectively (\cref{table:end2end}). To complement~\cref{fig:igc_vs_round}, the appendix includes~\cref{fig:num_instance_vs_round}, a plot of the distributions of the number of rounds for both methods.
\paragraph{Is there value in optimization-based separation?} 
\begin{figure}[ht]
    \centering
    \includegraphics[scale=0.25]{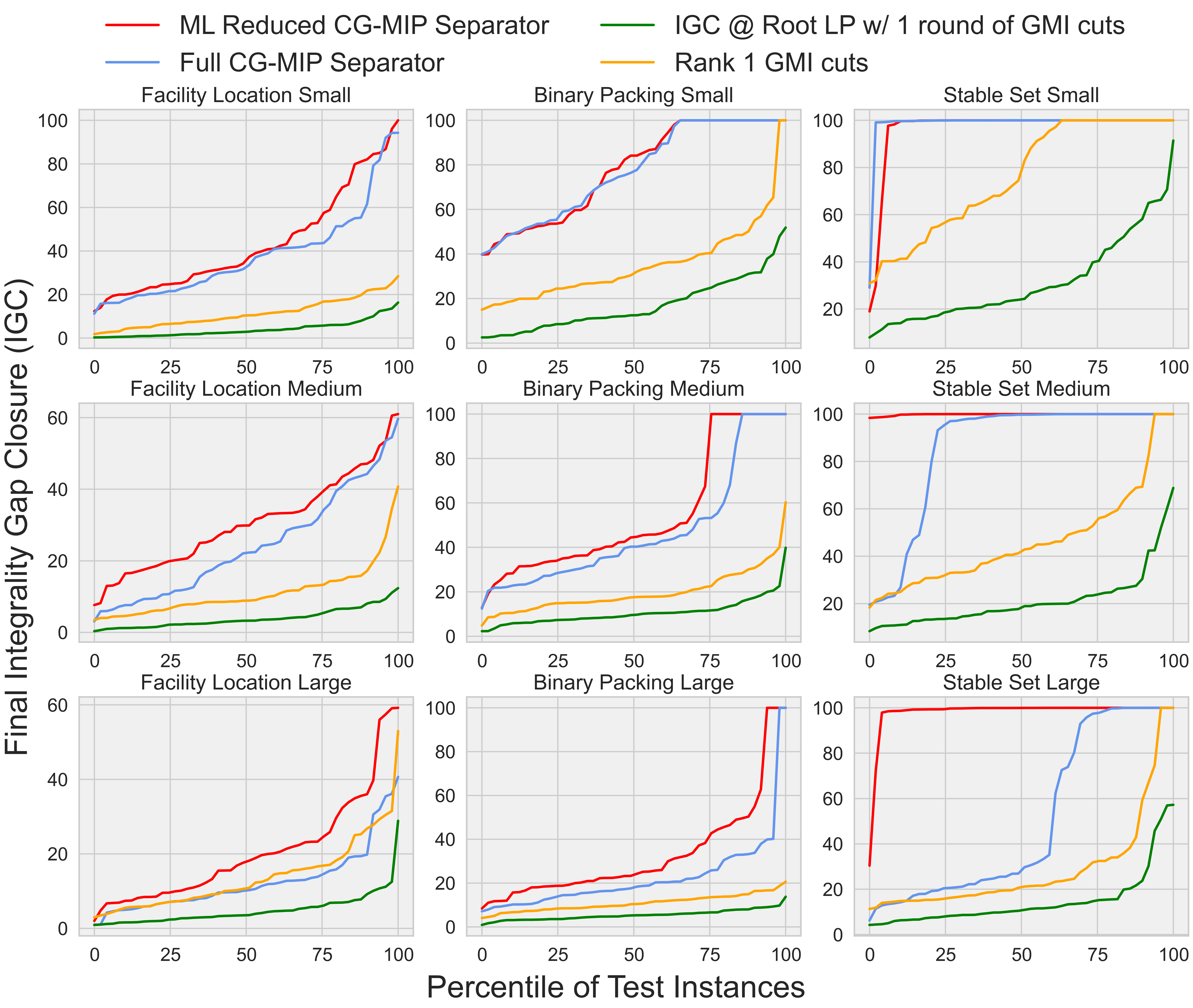}
    \caption{Plot of final test IGC v.s. percentile of instances. The reduced separator is shown in red, the full separator in blue, GMI cuts in yellow, and 1 round of GMI cuts in green. A larger area under a curve is preferred. Due to space limits, plots for SC and PM datasets are in Appendix~\cref{fig:igc_percentile_plot_PM_SC}.
    }
    \label{fig:IGC_percentile}
\end{figure}
To motivate the use of optimization-based separation, we compare both full and reduced CG-MIP to GMI cuts read from the simplex tableau in~\cref{fig:IGC_percentile}. The vertical axis corresponds to an IGC value and the horizontal axis corresponds to the percentile of test instances for which the final IGC is at most that. For most datasets, the reduced separator (in red) has the largest area under its curve, followed by the full separator (blue) and GMI cuts (yellow). This not only shows the benefit of the reduced separator over the full separator in terms of final IGC, but also the superiority of optimization-based cut generation over heuristic cut generation. The green curve represents the IGC obtained by adding one round of GMI cuts to the root LP relaxation at ``round zero'' of the cutting plane method, which all three methods use, and hence lower bounds the other curves.

\paragraph{Is there value in learning only from tight cuts?} 

\begin{figure}[t]
    \centering
    \includegraphics[scale=0.3]{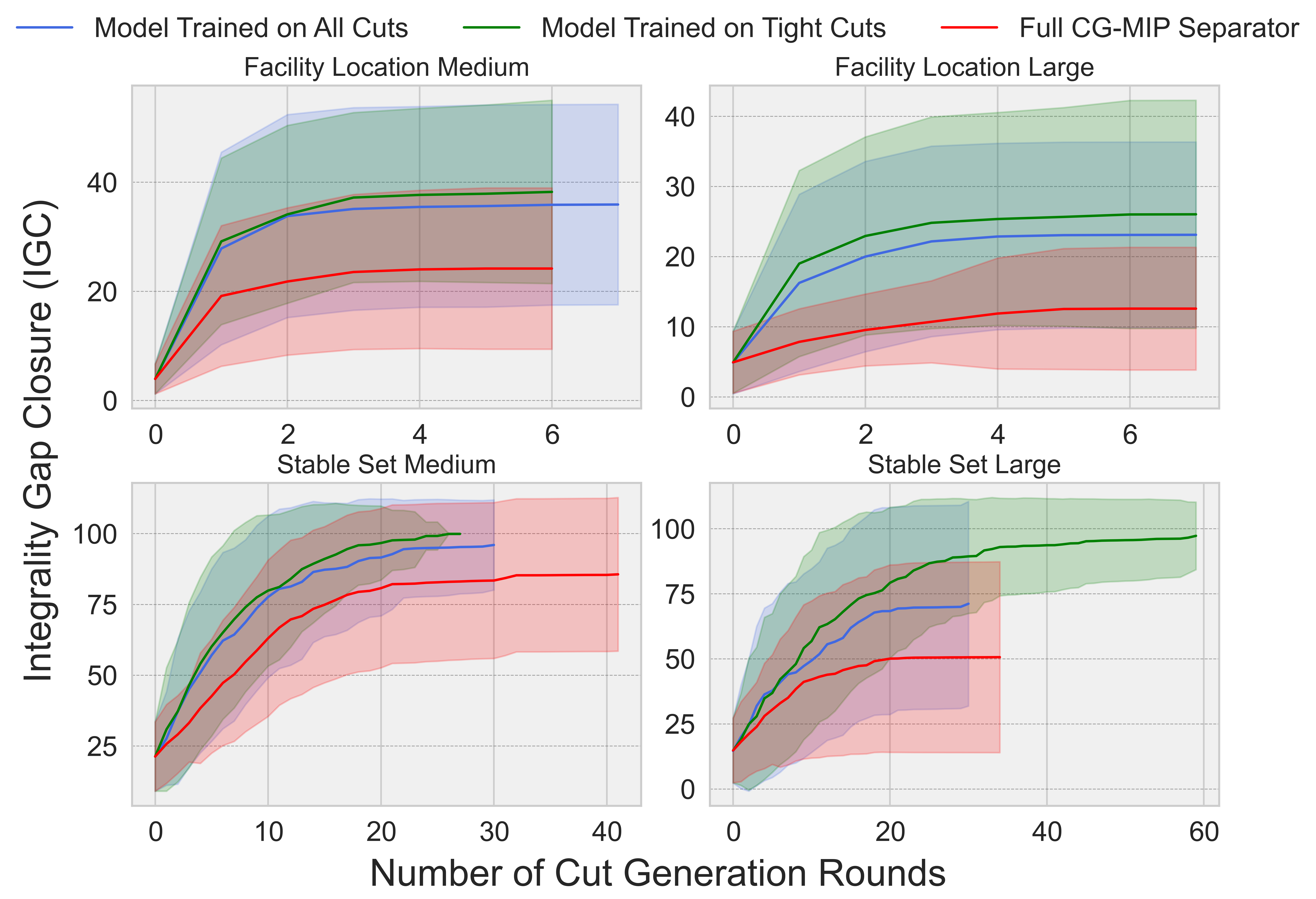}
    \caption{Plot of mean test IGC v.s. cut generation round for full and reduced separators with two labeling strategies.}
    \label{fig:igc_vs_round_CG_vs_tight}
\end{figure}

\begin{table}[t]
\centering
\scalebox{0.91}{
\begin{tabular}{cccccc}
\toprule
\multirow{2}{*}{\textbf{Dataset}} & \multirow{2}{*}{\textbf{Size}} & \textbf{Cut}  & \multicolumn{2}{c}{\textbf{Ratio}} & \textbf{\% Size} \\
 & & \textbf{Pool} & \textbf{IGC} $\uparrow$ & \textbf{Time} $\downarrow$ & \textbf{Reduction} \\
\midrule
\multirow{4}{*}{FL} & \multirow{2}{*}{M} & All & \textbf{1.54} & \textbf{0.84} & 96.03 \\
 & & Tight & \textbf{1.76} & \textbf{0.76} & 96.42 \\
\cmidrule{2-6} \
 & \multirow{2}{*}{L} & All & \textbf{1.95} & \textbf{0.62} & 96.59 \\
 & & Tight & \textbf{2.04} & \textbf{0.56} & 97.43 \\
\hline
\multirow{4}{*}{SS} & \multirow{2}{*}{M} & All & \textbf{1.21} & \textbf{0.46} & 61.29 \\
 & & Tight & \textbf{1.3} & \textbf{0.41} & 68.84 \\
\cmidrule{2-6} \
 & \multirow{2}{*}{L} & All & \textbf{1.42} & \textbf{0.79} & 65.12 \\
 & & Tight & \textbf{2.52} & 1.39 & 71.17 \\
\bottomrule
\end{tabular}
}
\caption{Ratio of mean test final IGC and time as cut pool being considered for classification labels varies.
}
\label{table:all_cuts_vs_tight}
\end{table}

As described in~\cref{sec:efficientcutselection}, the main variant of our method uses only tight (binding) cuts from CG-MIP's solution pool for labeling constraints for model training. As an ablation analysis, we compare classification models trained under both the aforementioned setting and the ``full'' setting in which all cuts are used.~\Cref{table:all_cuts_vs_tight} and~\cref{fig:igc_vs_round_CG_vs_tight} summarize the results. In both settings, the reduced CG-MIP separators dominate the full CG-MIP separator, as expected. However, we observe slightly better performance when learning from tight cuts. For example, in~\cref{fig:igc_vs_round_CG_vs_tight}, the green curves (tight cuts) consistently dominate both blue (all cuts) and red (baseline CG-MIP) curves. We believe this is due to judiciously ignoring some less-useful constraints and focusing on fewer, more impactful ones. This is also supported by the larger size reduction of CG-MIP and consequentially lower ratios of cut generation time in~\cref{table:all_cuts_vs_tight}.

\paragraph{Can we learn on small instances and generalize to larger ones?} 

\begin{table}[ht]
\centering
\scalebox{0.86}{
\begin{tabular}{cccccc} 
\toprule
\multirow{2}{*}{\textbf{Dataset}} & \textbf{Size}  & \multicolumn{2}{c}{\textbf{Ratio}} & \textbf{Abs. Perf} & \textbf{\% Size} \\
    &  \textbf{Test} & \textbf{IGC} $\uparrow$ & \textbf{Time} $\downarrow$ & \textbf{Win / Loss} & \textbf{Reduction} \\
\midrule
\multirow{2}{*}{SS}  & M &  \textbf{1.28} & \textbf{0.36}  & 74 / 0 & 73.04  \\
 &  L &  \textbf{2.65} & 1.3 & 34 / 4 & 77.57  \\
\hline
\multirow{2}{*}{PM}  & M &  \textbf{1.0} & \textbf{0.68} & 68 / 0 & 44.16  \\
 &  L &  \textbf{1.0} & \textbf{0.81}  & 60 / 4 & 31.74  \\
\hline
\multirow{2}{*}{SC}  & M &  \textbf{1.03} & \textbf{0.16}  & 76 / 0 & 77.61  \\
 &  L &  \textbf{1.14} & \textbf{0.1}  & 30 / 4 & 87.42  \\
\hline
\multirow{2}{*}{BP}  & M &  \textbf{1.04} & \textbf{0.36} & 66 / 4 & 97.92  \\
 & L &  \textbf{1.18} & \textbf{0.48} & 60 / 10 & 98.00  \\
\hline
\multirow{2}{*}{FL}  & M &  \textbf{1.44} & \textbf{0.7} & 34 / 2 & 94.45  \\
 &  L &  \textbf{1.92} & \textbf{0.51} & 54 / 0 & 98.03  \\
\bottomrule
\end{tabular}%
    }
\caption{End-to-end metrics of reduced separator using GNN trained on small instances evaluated on larger instances.}
\label{table:generalization}
\end{table}

\cref{table:generalization} shows how models trained on small instances fare on medium and large instances. In most cases, we observe a substantial size reduction of CG-MIP while maintaining or improving the final IGC in significantly less time.~\Cref{fig:boxplot_time_generalization} in the appendix illustrates the logarithmic total time comparison via boxplots, revealing that the reduced separator generally achieves a lower total time distribution in most datasets, highlighting the computational efficiency of the reduced CG-MIP separator.

\section{Discussion and Conclusion}\label{sec:ch8}
This paper presents a framework for integrating ML into an established MILP framework for optimization-based separation of aggregation-based cuts. We reduce the computational burden of exact separators by identifying a small important subset of constraints to consider for aggregation. Although we focus on CG cuts, our framework can be applied to other families of aggregation-based cuts. The framework relies on the use of graph neural networks trained on a supervised binary classification task to identify constraints that lead to the generation of strong CG cuts. Extensive computational experiments demonstrate that the trained models can effectively reduce the size of the separation problem with little effect on cut quality and smaller runtimes across various MILPs. Models generalize well to unseen data and scale effectively to larger problem sizes. 

Although our approach demonstrates faster solution times and improved final IGC, several limitations and future directions remain. First, our evaluation focuses on a pure cutting plane method, not integrated within a MILP solver, which poses challenges due to stricter time constraints for cut generation. For instance, SCIP's default CG-MIP setting (although disabled by default) is to halt at the first incumbent solution, and hence may challenge the competitive edge of our approach. Second, the use of supervised learning for classification could be enhanced by exploring reinforcement or imitation learning, which may better capture the iterative nature of the cutting plane method. Additionally, to enhance model robustness and diversity of LP solutions to separate, we can augment data points by perturbing LP solutions, thereby increasing dataset size with minimal cost. Finally, to further reduce CG-MIP's size, we can learn to predict upper bounds on $\boldsymbol{u}$ through regression rather than classifying constraints. This could further reduce CG-MIP's search space, as initial findings indicate that many coefficient values are often very small nonnegative numbers. However, preliminary results also reflect a common challenge in ML: regression tasks tend to be more difficult than binary classification.

\clearpage
\bibliography{aaai25}

\clearpage
\appendix

\section{Projected CG Cuts}
\label{app:procg}
CG-MIP applies only to pure ILP but can be extended to MILP by the projected Chv\'atal-Gomory cuts of~\citet{Bonami2008}. A projected CG cut is an inequality satisfying:
\begin{equation}\label{eq:procg}
\begin{aligned}
    \lfloor \boldsymbol{u}^T \boldsymbol{A} \rfloor \boldsymbol{x} \leq  \lfloor \boldsymbol{u}^T\boldsymbol{b}\rfloor & 
     \ \ \ \forall \  \boldsymbol{u}\in\mathbb{R}^{m}_+ \ \ \text{s.t. } \ \boldsymbol{u}^T \boldsymbol{C} \geq \boldsymbol{0}, \\
\end{aligned}
\end{equation}
where the matrix $\boldsymbol{C} \in \mathbb{Q}^{m \times p}$ corresponds to the constraint matrix of the $p$ continuous variables. CG-MIP can be extended to generate projected CG cuts by adding constraint~\eqref{eq:procg} to problem~\eqref{eq:cgmip}. 

\section{Benchmark Problems}
\label{sec:synthetic_MILP_instances}
\paragraph{Binary Packing} This problem follows the general ILP formulation in equation~\eqref{eqn:ILP} but requires the coefficients in $A$, $b$, and $c$ to be nonnegative as well as binary decision variables. In terms of instance generation, we follow the methodology used in \cite{Columbia}, where entries of $A$ are integers uniformly sampled in the range $(5,30)$. Each entry in $c$ is integers uniformly sampled in the range $(1,10)$. Finally, each entry in $b$ is integers sampled uniformly in the range $(10n,20n)$.

\begin{equation*}
\begin{aligned}
    \min_{\mathbf{x}\in\{0,1\}^{n}} \ & \mathbf{c}^\intercal \mathbf{x} \\
    \text{s.t.} \ & \mathbf{A}\mathbf{x} \leq \mathbf{b}\\
    & {\mathbf{x}\in\{0,1\}^{n}}
\end{aligned}
\end{equation*}

\paragraph{Capacitated P-median} This extension of the classic p-median problem incorporates facility capacities. Given a set of customers \( I = \{1, \ldots, n\} \) with demands \( d_i \), and facilities with capacities \( c_j \) and costs \( w_{ij} \) for serving customer \( i \) from facility \( j \), the goal is to minimize total service cost while ensuring that no facility's capacity is exceeded. Binary variables \( y_j \) and \( x_{ij} \) indicate if a facility is opened and if customer \( i \) is assigned to facility \( j \), respectively. We use MIPLearn \cite{miplearn} for instance generation with their default parameters. The problem is formulated as:

\begin{align*}
\min   \quad & \sum_{i \in I} \sum_{j \in I} w_{ij} x_{ij} &\\
\text{subject to} \quad & \sum_{j \in I} x_{ij} = 1 & \quad \forall i \in I \\
                        & \sum_{j \in I} y_{j} = p &\\
                        & \sum_{i \in I} d_i x_{ij} \leq c_j y_{j} & \quad \forall j \in I \\
                        & x_{ij} \in \{0,1\} & \quad \forall i,j \in I \\
                        & y_{j} \in \{0,1\} & \quad \forall j \in I
\end{align*}
The constraints ensure each customer is assigned to exactly one facility, exactly \( p \) facilities are opened, and no facility exceeds its capacity.

\paragraph{Set-Cover} 
The set cover problem aims to find the minimum number of sets that cover all elements in a given universe. The binary linear program formulation is:

\begin{align*}
\min & \sum_{s \in S} x_s \\
\text{subject to} 
& \sum_{s \in S: e \in s} x_s \geq 1 & \forall e \in U \\
& x_s \in \{0, 1\} & \forall s \in S.
\end{align*}
Here, $x_s$ is a binary variable indicating whether set $s$ is selected. The constraint ensures every element in the universe $U$ is covered by at least one selected set. We use \cite{prouvost2020ecolegymlikelibrarymachine} for instance generation with their default instance parameters.

\paragraph{Stable Set} The maximum-weight stable set problem seeks the heaviest subset of vertices in a graph where no 2 vertices are adjacent. Given a simple undirected graph $ G = (V, E)$ with vertex weights $w_v$, the problem is formulated as:

\begin{align*}
\min \quad & - \sum_{v \in V} w_v x_v &\\
\text{subject to} \quad & x_v + x_u \leq 1 &\quad \forall (v, u) \in E \\
                       & x_v \in \{0, 1\} &\quad \forall v \in V
\end{align*}
The objective is to maximize the total weight of selected vertices by minimizing the negative weight sum. Constraints ensure no two adjacent vertices are selected, maintaining the set's stability. The binary variable \( x_v \) indicates whether vertex \( v \) is in the stable set. We use MIPLearn \cite{miplearn} for instance generation with their default parameters.

\paragraph{Capacitated Facility Location} This problem involves optimizing the placement and operation of facilities to serve customers at minimal cost, considering both fixed facility and transportation costs. Each facility has a limited capacity, and the goal is to allocate customer demand without exceeding these capacities. In the variant used here, customers can be served by multiple facilities with continuous allocation variables. Following the instance generation by Cornuejols et al. (1991), let \( I = \{1, \ldots, n\} \) be the customers and \( J = \{1, \ldots, m\} \) the facilities. For each customer \( i \), \( d_i \) represents demand, and for each facility \( j \), \( c_j \) denotes capacity, \( f_j \) the fixed cost, and \( t_{ij} \) the transportation cost. The decision variables are \( x_{ij} \) (continuous, indicating the fraction of customer \( i \)'s demand that is served by facility \( j \)) and \( y_j \) (binary, indicating if facility \( j \) is opened).

\begin{align*}
\min & \sum_{j \in J} f_j y_j + \sum_{i \in I} \sum_{j \in J} t_{ij} x_{ij}& \\
\text{subject to}
& \sum_{j \in J} x_{ij} = 1 &\quad \forall i \in I \\
& \sum_{i \in I} d_i x_{ij} \leq c_j y_j &\quad \forall j \in J \\
& \sum_{j \in J} c_j y_j \geq \sum_{i \in I} d_i \\
& x_{ij} \leq y_j &\quad \forall i \in I, \forall j \in J \\
& x_{ij} \in [0, 1] &\quad \forall i \in I, \forall j \in J \\
& y_j \in \{0, 1\} &\quad \forall j \in J
\end{align*}
The first constraint ensures each customer is served by one facility. The second ensures facility capacity is not exceeded. The third guarantees total capacity meets total demand, and the fourth ensures customers are only assigned to open facilities. We use \cite{prouvost2020ecolegymlikelibrarymachine} for instance generation.

\section{Variable and Constraint Features}
\label{sec:data_feats}

\cref{tab:feature_description} describes the variable, constraint and edge features that we use in our VCG representation of state $S_t$ in the cutting plane mehtod. VCG is represented by the tuple $(\mathcal{C}, \mathcal{E}, \mathcal{V})$ where $\mathcal{C} \in \mathbb{R}^{m \times 53}$ represents the constraint node feature matrix, $\mathcal{V} \in \mathbb{R}^{n \times 18}$ represents the variable node feature matrix and $\mathcal{E} \in \mathbb{R}^{(m \cdot n) \times 1}$ represent the edge feature vector.

\section{Implementation Details}
\label{app:impl_eval_details}

\paragraph{Baseline implementation and parameters}
The implementation of the GMI cuts is that of the SCIP solver\footnote{\url{https://github.com/scipopt/PySCIPOpt/blob/master/tests/test_gomory.py}.}. The \textit{emphasis} parameter is documented in the SCIP documentation\footnote{See \url{https://www.scipopt.org/doc/html/group__ParameterMethods.php#gab2bc4ccd8d9797f1e1b2d7aaefa6500e}.}. This ensures that SCIP will spend more resources earlier on in the search on primal heuristics to find feasible solutions fast. The cutting plane loop will terminate early if the IGC does not improve by 0.001 over 7 rounds to avoid stagnation and diminishing returns. As done in \cite{lodi}, CG-MIP's solution pool is filtered to discard CG cuts with the same violation, except the one with the sparsest support. Additionally, we also push CG-MIP to prioritize the generation of CG cuts with sparse aggregation coefficients by introducing the penalty term $-\sum_{i=1}^{m}w_iu_i$ to CG-MIP's objective funciton, where $w_i=1e^{-4}$ for all $i$.

\paragraph{Shifted geometric mean} The shifted geometric mean of a set of $n$ values $t_1, \dots, t_n$ is defined as $\left( \prod_{i=1}^{n} [t_i + \text{shift}] \right)^{\frac{1}{n}} - \text{shift}$. Compared to the arithmetic mean, it is less sensitive to large variations in the values~\cite{Tobias_thesis}.

\paragraph{Training} Hyper-parameter tuning is done using 80 trials of random search. 
~\cref{table:hyperparams1} lists all hyperparameters and their domains; ~\cref{table:hyperparams2} lists the final tuned values for each problem along with the optimized classification threshold. All GNN models are trained for 700 epochs using the Adam optimizer with a learning rate scheduler (ReduceLROnPlateau) to adjust the learning rate when progress stalls. Early stopping is used to terminate training if the validation performance does not improve within a set number of epochs.
\paragraph{Software} All experiments were run on a computing cluster with an Intel Xeon CPU E5-2683 and Nvidia Tesla P100 GPU with 64GB of RAM (for training). SCIP 8.0.0 \cite{scip8} was used as the MIP solver. Pytorch 1.10 \cite{paszke2019pytorchimperativestylehighperformance} was used for supervised learning.

\section{Additional Experimental Results}

\begin{table}[H]
\centering
\scalebox{0.99999}{
\begin{tabular}{cccc} 
\hline
\textbf{Dataset} & \textbf{Size} & \textbf{Test Accuracy} & \textbf{Test Recall} \\
\hline
\multirow{3}{*}{SS} & S & 79.40 & 66.33 \\
 & M & 87.33  & 57.15  \\
 & L & 88.19 & 56.82 \\
\hline
\multirow{3}{*}{PM} & S & 85.54 & 91.02 \\
 & M & 76.42 & 88.64  \\
 & L & 70.03 & 91.32 \\
\hline
\multirow{3}{*}{SC} & S  & 77.44 & 82.07  \\
 & M  & 81.50  & 61.62 \\
 & L  & 91.33  & 40.79 \\
\hline
\multirow{3}{*}{BP} & S  & 98.24 & 92.90 \\
 & M  & 99.19  & 92.60 \\
 & L & 99.35 & 92.63 \\
\hline
\multirow{3}{*}{FL} & S  & 96.39  & 57.95 \\
 & M  & 97.44  & 72.31  \\
 & L  & 98.48  & 69.09 \\
\hline
\end{tabular}
}
\caption{Test Accuracy and recall of trained GNN's}
\label{table:classification_results}

\end{table}

\begin{table}[ht]
\centering
\scalebox{0.9999}{
\begin{tabular}{cccccc} 
\hline
\multirow{2}{*}{\textbf{Dataset}} & \multirow{2}{*}{\textbf{Size}} & \multicolumn{2}{c}{\textbf{Mean IGC}} & \multicolumn{2}{c}{\textbf{Mean Time}} \\
    &  & \textbf{Reduced} & \textbf{Full} & \textbf{Reduced} & \textbf{Full}  \\
\hline
\multirow{3}{*}{SS} & S & 93.56 & 97.49 & 40.92 & 122.91 \\
 & M & 99.86 & 78.14 & 69.07 & 346.27 \\
 & L & 96.76 & 37.39 & 137.82 & 153.89 \\
\hline
\multirow{3}{*}{PM} & S & 99.81 & 99.98 & 5.91 & 19.88 \\
 & M & 99.38 & 99.63 & 30.65 & 54.99 \\
 & L & 99.65 & 99.98 & 62.19 & 94.55 \\
\hline
\multirow{3}{*}{SC} & S & 96.97 & 99.87 & 17.65 & 37.94 \\
 & M & 97.60 & 74.32 & 76.51 & 318.32 \\
 & L & 89.84 & 30.03 & 203.75 & 308.74 \\
\hline
\multirow{3}{*}{BP} & S & 74.46 & 74.10 & 61.71 & 131.37 \\
 & M & 48.25 & 41.29 & 222.93 & 678.61 \\
 & L & 27.60 & 19.14 & 207.37 & 570.63 \\
\hline
\multirow{3}{*}{FL} & S & 37.88 & 33.33 & 23.28 & 43.00 \\
 & M & 27.67 & 19.57 & 24.34 & 47.82 \\
 & L & 16.54 & 10.18 & 22.91 & 58.02 \\
\hline
\end{tabular}
}
\caption{Raw mean final IGC and cumulative cut generation time (in seconds) for test instances for the ML-driven reduced CG-MIP and full exact CG-MIP.}
\label{tab:raw_end2end}
\end{table}

\begin{figure}[h]
    \centering
    \includegraphics[scale=0.27]{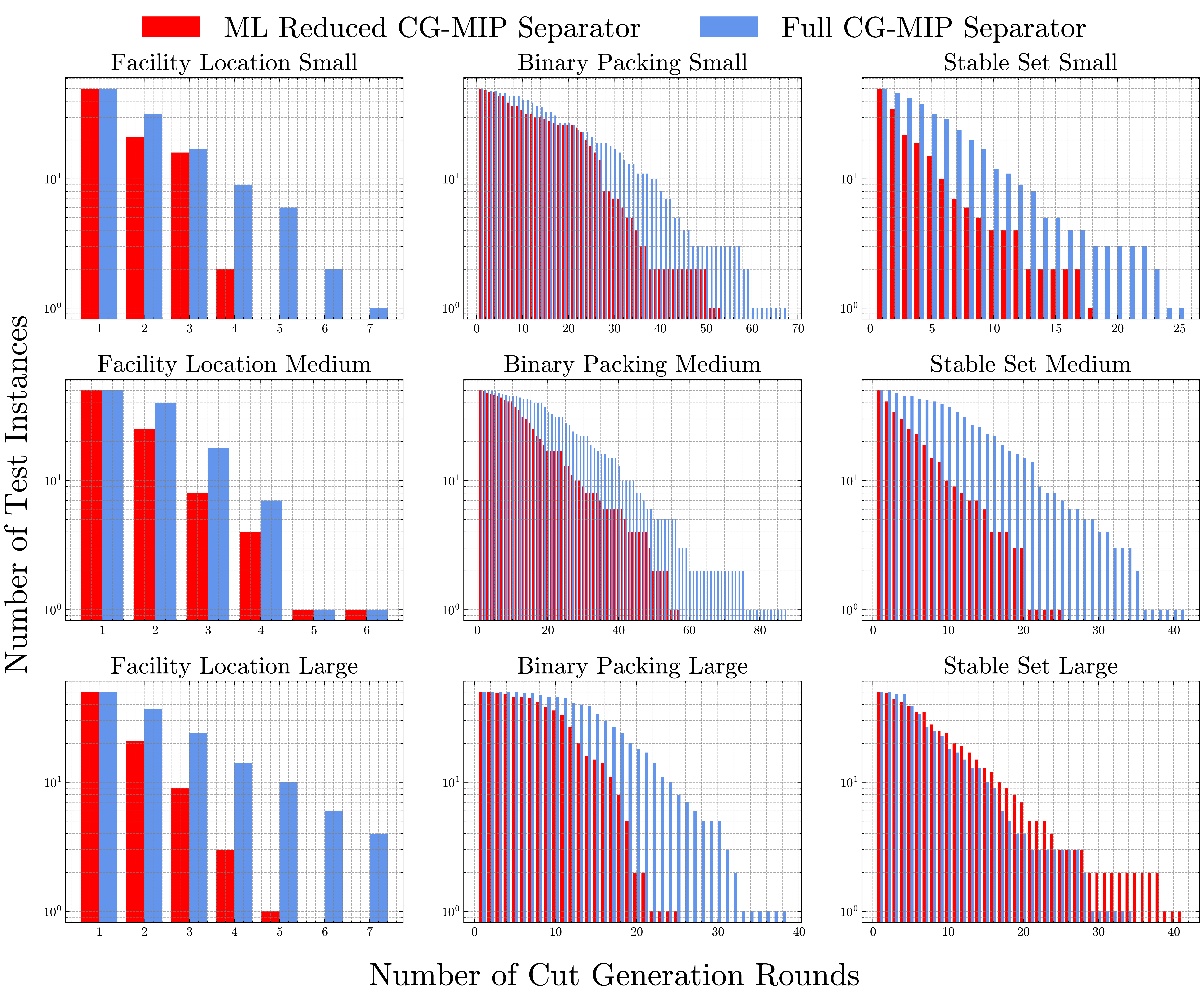}
    
    \caption{Plot of number of test instances that made it to a specific round of the cutting plane method for both reduced and full CG-MIP separator.}
    \label{fig:num_instance_vs_round}
\end{figure}

\begin{figure}[h]
    \centering
    
    \includegraphics[scale=0.31]{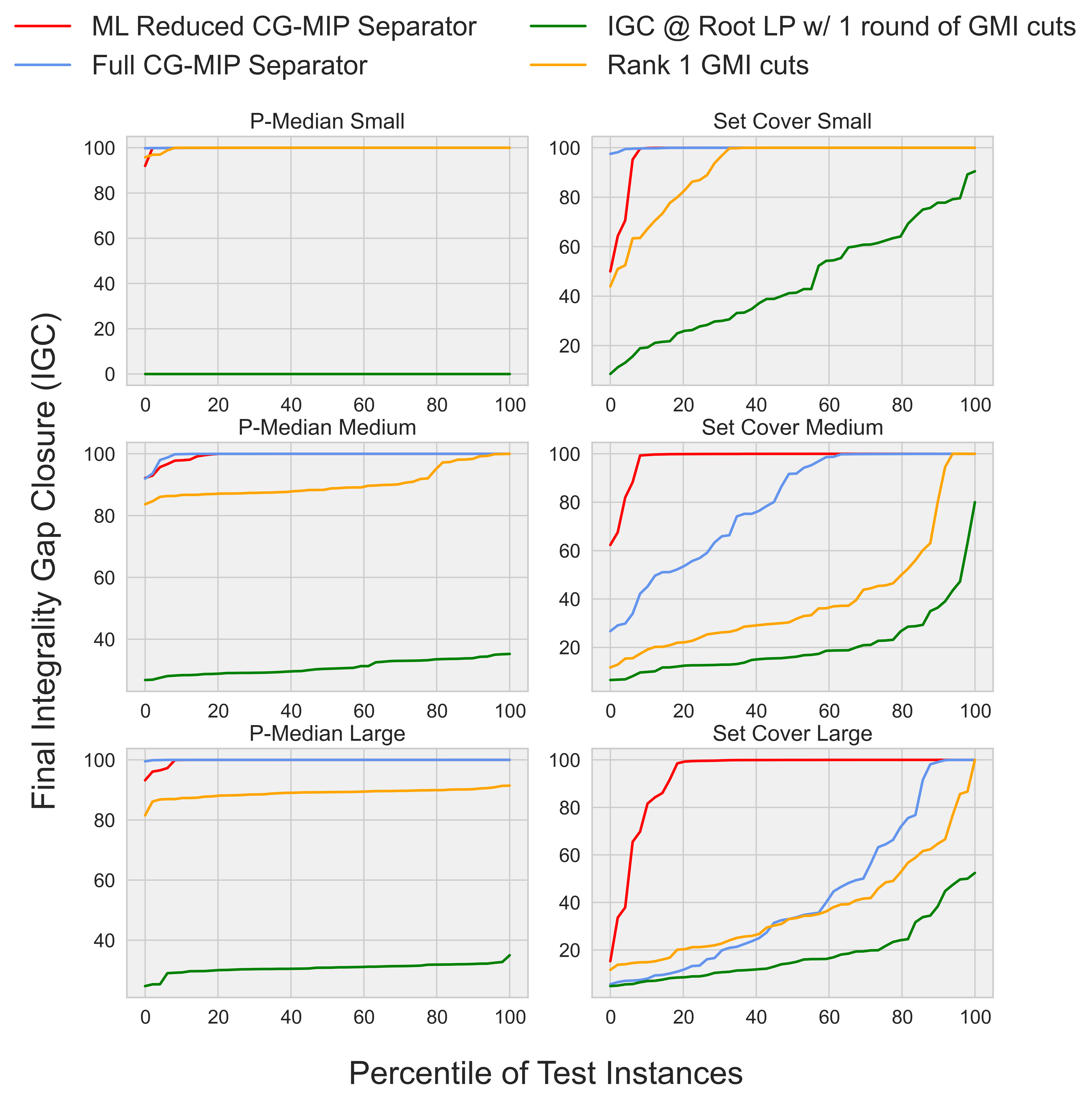}
    \caption{Plot of final test IGC v.s. percentile of instances for P-Median and Set Cover datasets. The reduced separator is shown in red, the full separator in blue, GMI cuts in yellow, and 1 round of GMI cuts in green. A larger area under a curve is preferred.}
    \label{fig:igc_percentile_plot_PM_SC}
\end{figure}

\begin{figure}[h]
    \centering
    \includegraphics[scale=0.29]{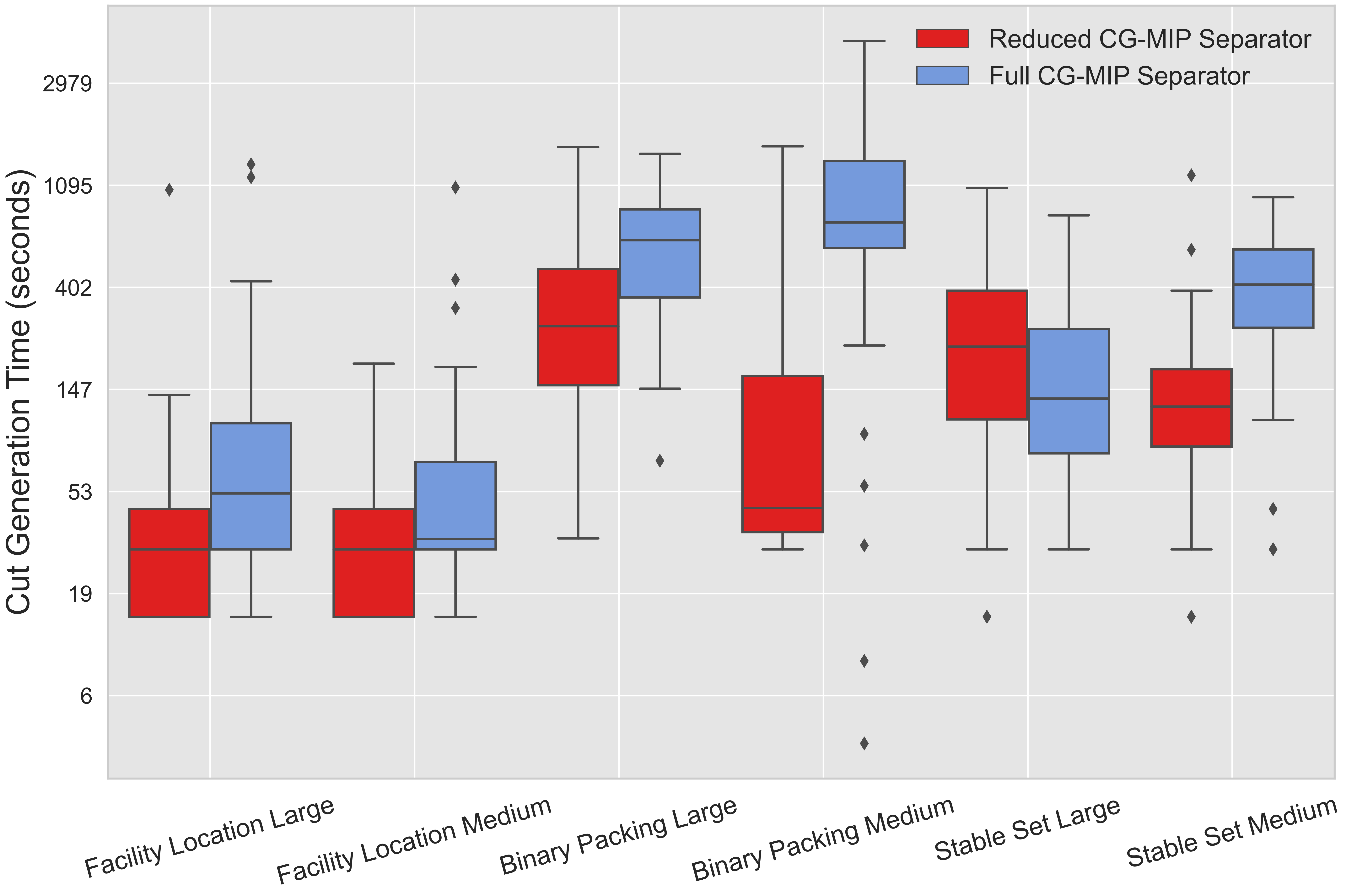}
    
    \caption{Boxplot of cut generation time for reduced and full separator when generalizing to medium/large instances.}
    \label{fig:boxplot_time_generalization}
\end{figure}

\begin{table*}[b]
\centering
\begin{tabular}{@{}cll@{}}
\toprule
Tensor & Feature       & Description \\
\midrule                
$\boldsymbol{C}$       & obj\_cos\_sim & Cosine similarity with objective. \\
        & bias          & Bias value, normalized with constraint coefficients. \\
        & is\_tight     & Tightness indicator in LP solution. \\
        & dualsol\_val  & Dual solution value, normalized. \\
        & row\_at\_lbs           & Lower bound indicator. \\
        & row\_at\_ubs           & Upper bound indicator. \\
        & norm\_slack           & slack of constraint normalized. \\
        & norm\_violation           & relative violation \\
        & euclidean\_distance           & distance of constraint to fractional solution \\
        & type           & linear, logicor, knapsack, setppc, varbound as one-hot encoding. \\
        & constraint\_degree           & number of active variables at constraint \\
        & zero\_rhs           & Indicator of rhs being 0. \\
        & A           & mean, median, std, min, max of all constraint coefficients \\
        & A\_ratio           & max, min, mean, std of ratio of all constraint coefficients \\
        & A\_{nonzero}           & mean, median, std, min, max of constraint coefficients when vars $>$ 0\\
        & A\_ratio\_{nonzero}           & max, min, mean, std of ratio of constraint coefficients when vars $>$ 0\\
        & A\_{zero}           & mean, median, std, min, max of constraint coefficients when vars $=$ 0\\
        & A\_ratio\_{zero}           & max, min, mean, std of ratio of constraint coefficients when vars $=$ 0\\
        & A\_{at\_UB}           & mean, median, std, min, max of constraint coefficients when vars at upper bound\\
        & A\_ratio\_{at\_UB}           & max, min, mean, std of ratio of constraint coefficients when vars at upper bound\\
        
\midrule
$\boldsymbol{E}$      & coef          & Constraint coefficient, normalized per constraint. \\
        
\midrule
$\boldsymbol{V}$       & type          & Type (binary, integer, impl. integer, continuous) as a one-hot encoding. \\
        & coef              & Objective coefficient, normalized. \\
        & has\_lb           & Lower bound indicator. \\
        & has\_ub           & Upper bound indicator. \\
        & lb                & Lower bound. \\
        & ub                & Upper bound. \\
        & sol\_is\_at\_lb   & Solution value equals lower bound. \\
        & sol\_is\_at\_ub   & Solution value equals upper bound. \\
        & sol\_frac         & Solution value fractionality. \\
        & basis\_status     & Simplex basis status (lower, basic, upper, zero) as a one-hot encoding. \\
        & reduced\_cost     & Reduced cost, normalized. \\
        & sol\_val          & Solution value normalized. \\
        
\bottomrule
\end{tabular}
\caption{Description of the constraint, edge and variable features in our VCG state representation $(\mathcal{C}, \mathcal{E}, \mathcal{V})$.}
\label{tab:feature_description}

\end{table*}

\begin{table*}[h!]
\centering
\begin{tabular}{c|c|c|c|c|c}
\hline
\textbf{Hyperparameter Name} & \textit{Learning Rate} & \textit{Weight Decay} & \textit{Dropout Rate} & \textit{Embedding Size} & \textit{Batch Size}\\ \hline
\textbf{Domain} & $\left[ 1\text{e-5},\ 1\text{e-1} \right]$ & $\left[ 1\text{e-5},\ 1\text{e-1} \right]$ & $\left[0,0.5\right]$ & $\{4,8,16,32,64\}$ & $\{16,32,64,128\}$ \\ \hline
\end{tabular}
\caption{Hyperparameter space for GNN models. All hyperparameters are sampled independently and uniformly at random from their respective domains.}
\label{table:hyperparams1}

\end{table*}

\begin{table*}[h!]
\centering
\resizebox{\textwidth}{!}{
\begin{tabular}{c|ccc|ccc|ccc|ccc|ccc}
\hline
\multirow{2}{*}{\textbf{Hyperparameter Name}} & \multicolumn{3}{c|}{\textbf{SS}} & \multicolumn{3}{c|}{\textbf{SC}} & \multicolumn{3}{c|}{\textbf{FL}} & \multicolumn{3}{c|}{\textbf{BP}} & \multicolumn{3}{c}{\textbf{PM}} \\ \cmidrule{2-16}
 &  \textbf{S} & \textbf{M} & \textbf{L} & \textbf{S} & \textbf{M} & \textbf{L} & \textbf{S} & \textbf{M} & \textbf{L} & \textbf{S} & \textbf{M} & \textbf{L} & \textbf{S} & \textbf{M} & \textbf{L} \\ \hline
\textit{Learning Rate} & 0.01263 & 0.02286 & 0.02361 & 0.00931 & 0.02467 & 0.01095 & 0.00827 & 0.00645 & 0.00468 & 0.02365 & 0.01091 & 0.00564 & 0.00312 & 0.00569 & 0.00134 \\ \hline
\textit{Weight Decay} & 0.01573.0 & 0.0 & 0.02154 & 0.09124 & 0.0 & 0.0 & 0.0 & 0.00057 & 0.05871 & 0.0 & 0.0 & 0.0 & 0.0 & 0.0 & 0.0 \\ \hline
\textit{Dropout Rate} & 0.03012 & 0.0182 & 0.08233 & 0.02105 & 0.12415 & 0.05117 & 0.09301 & 0.08136 & 0.10431 & 0.13532 & 0.02085 & 0.03293 & 0.00417 & 0.02685 & 0.03689 \\ \hline
\textit{Embedding Size} & 64 & 64 & 32 & 64 & 64 & 64 & 8 & 16 & 32 & 32 & 32 & 32 & 32 & 32 & 32 \\ \hline
\textit{Batch Size} & 64 & 64 & 64 & 32 & 64 & 64 & 32 & 32 & 32 & 64 & 32 & 32 & 64 & 64 & 64 \\ \hline \hline
\textit{Classification Threshold} & 0.65 & 0.77 & 0.77 & 0.64 & 0.72 & 0.77 & 0.79 & 0.93 & 0.96 & 0.91 & 0.95 & 0.98 & 0.6 & 0.47 & 0.40 \\ \hline

\end{tabular}
}
\caption{Final tuned values for all hyperparameters.}
\label{table:hyperparams2}

\end{table*}

\end{document}